\documentclass[10pt,twocolumn,letterpaper]{article}

\usepackage{iccv}
\usepackage{times}
\usepackage{epsfig}
\usepackage{graphicx}
\usepackage{amsmath}
\usepackage{amssymb}
\usepackage{paralist}

\usepackage{algorithm}
\usepackage{algorithmic}
\usepackage{xcolor}

\newcommand{\abc}[1]{\textcolor{black}{#1}}
\newcommand{\abcn}[1]{\textcolor{black}{#1}}
\newcommand{\yty}[1]{\textcolor{black}{#1}}
\newcommand{\ytyn}[1]{\textcolor{black}{#1}}
\newcommand{\comments}[1]{}

\newcommand{\argmax}{\mathop{\mathrm{argmax}}}

\usepackage[pagebackref=true,breaklinks=true,letterpaper=true,colorlinks,bookmarks=false]{hyperref}

\iccvfinalcopy 

\def\iccvPaperID{2183} 

\ificcvfinal\fi
\begin{document}

\title{Recurrent Filter Learning for Visual Tracking}

\author{Tianyu Yang \hspace{0.8cm} Antoni B. Chan\\
City University of Hong Kong\\
{\tt\small tianyyang8-c@my.cityu.edu.hk, abchan@cityu.edu.hk}
}

\maketitle

\begin{abstract}
\abc{
   Recently using convolutional neural networks (CNNs) has gained popularity in visual tracking, due to its robust feature representation of images.
   Recent methods perform online tracking by fine-tuning a pre-trained CNN model to the specific target object using
stochastic gradient descent (SGD) back-propagation, which 
is usually time-consuming.
   %
In this paper, we propose a recurrent filter generation methods for visual tracking.  We directly feed the target's image patch to a recurrent neural network (RNN) to estimate an object-specific filter for tracking. 
As the video sequence is a spatiotemporal data, we extend the matrix multiplications of the fully-connected layers of the RNN to a convolution operation on feature maps, 
which preserves the target's spatial structure and also is memory-efficient. 
The tracked object in the subsequent frames will be fed into the RNN to adapt the generated filters to appearance variations of the target. 
Note that once the off-line training process of our network is finished, there is no need to fine-tune the network for specific objects, which makes our approach more efficient than methods that use iterative fine-tuning to online learn the target.
Extensive experiments conducted on widely used benchmarks, OTB and VOT,  demonstrate encouraging results compared to other recent methods.
}
\end{abstract}

\section{Introduction}

Given an object of interest labeled by a bounding box in the first frame, the goal of generic visual tracking is to locate this target in subsequent frames automatically. Since the \abc{type of} object may not be known in advance under some scenarios, 
\abc{it is infeasible} 
to gather a lot of data to train an \abc{object-}specific tracker. 
\abc{Hence, generic visual tracking should be robust enough to work with any type of object, while also being sufficiently adaptable to handle the appearance of the specific object and variations of appearance during tracking.}
Visual tracking 
plays a crucial role in numerous vision application such smart surveillance, autonomous driving and human-computer interaction.

Most existing algorithms locate the object by online training a discriminative model to classify the target from the background. This self-updating paradigm assumes 
that the object's appearance 
changes smoothly, but is 
inappropriate in 
challenging situations such as 
heavy occlusion, illumination changes and abrupt motion. Several methods adopt multiple experts \cite{Zhang2014}, multiple instance learning \cite{Babenko2011}, or \ytyn{short and long term memory stores}
\cite{Hong2015} to address 
the problem of drastic appearance changes. 
Recent advances using CNNs for object recognition and detection has inspired 
tracking algorithms to employ 
the \abc{discriminative} 
features learned by CNNs. 
In particular, 
\cite{Ma2015, Qi2016, Danelljan2015} feed the CNN features into 
\abc{a traditional visual tracker}, 
the 
correlation filter \cite{Henriques2015}, to get a response map for target's estimated location. 

\abc{However, the application domain of object recognition/detection is quite different from visual tracking.
In object recognition, the networks are trained to recognize specific categories of objects, whereas in visual tracking the type of object is unknown, and varies from sequence to sequence.  Furthermore, because the  CNNs are trained to recognize object classes (e.g., dogs), the high-level features are {\em invariant} to appearance variations within the object class (e.g., a brown dog vs. a white dog).  In contrast, visual tracking is concerned about recognizing a {\em specific instance} of an object (a brown dog), possibly among distractor objects in the same class (white dogs).}
Thus, \ytyn{naively} applying CNNs trained on 
the object recognition task is not suitable. 
One way to 
address this problem is to 
\abc{fine-tune} a pre-trained CNN models for each test sequence \abc{from} 
the first frame, but this is 
time-consuming and is prone to overfitting due to the limited available labeled data. 
Therefore, using a smaller learning rate, \ytyn{constraining} the number of iteration for SGD back-propagation, and only fine-tuning fully-connected layers \cite{Nam2016, Nam20161} 
\abc{have been proposed for adapting the CNN to the specific target, while also}
alleviating the risk of ruining the weights of model.

In contrast to these
object-specific methods, we propose a {\em recurrent filter learning} (RFL) algorithm by maintaining the target appearance \abc{and tracking filter} through a Long Short Term Memory (LSTM) network. A fully convolutional neural networks is used to encode the target appearance information, while preserving the spatial structure of the target. 
\abc{Naively flattening the CNN feature map into a vector in order to pass it to an LSTM would obfuscate the structure of the target.}
Instead, inspired by \cite{Shi2015}, we 
\abc{change the input, output, cell and hidden states of the LSTM into feature maps, and use convolution layers instead of fully-connected layers.}
\abc{The output of the convolutional-LSTM is itself a filter, which is then convolved with the feature map of the next frame to produce the target response map.} 

Our RFL network has several interesting properties. First, once the offline training process is finished, there is no need to retrain the network for specific objects during test time. 
\abc{Instead, a purely feed-forward process is used, where, for each frame, the image patch of the tracked target is input into the network, which then updates its appearance model and produces a target-specific filter for finding the target in the next frame.}
\ytyn{Second, the object does not have to be at the center of exemplar patch due the \abcn{fully} convolutional structure of our LSTM, which is shift invariant. This is very useful for updating the tracker, since the estimated bounding box for each frame may not be accurate, \abcn{resulting in shifting or stretching the target in the exemplar image}.}
Furthermore, 
the convolution of the search image and the generated filter is actually a dense sliding window searching strategy, \abcn{implemented in a more efficient way.}
%
\yty{The contribution of this paper are three-fold:}

\begin{compactenum}
	\item We propose a novel recurrent filter learning framework that captures both spatial and temporal information of sequence for visual tracking, \abc{and does not require fine-tuning during tracking}.
	
	\item We design an 
	efficient and effective method for initializing and updating 
	the target appearance 
	for a specific object by using a convolutional LSTM as the memory unit.
	
	\item We conduct extensive experiments on the widely used benchmark OTB \cite{Wu2015} and VOT \cite{Kristan2016}, which contains 100 challenging sequences and 60 videos respectively to evaluate the effectiveness of our tracker. 
\end{compactenum}


\section{Related Work}
\subsection{CNN-based Trackers}

The past several years have seen great success of CNNs on various computer vision tasks, especially on object recognition and detection\cite{Simonyan2015, He2016, Ren2015, Redmon2016}. Recently, several 
trackers embed CNNs into their frameworks due to its outstanding \abc{power of its feature representations.} 
One straightforward way of using CNNs is treating it as a feature extractor. 
\cite{Ma2015} adopts the output of some specific layers of VGG-net \cite{Simonyan2015} including high-level layers, which encodes the semantic information, and low-level layers which preserve the spatial details. 
\cite{Danelljan2015} uses a similar approach 
by using the activation of the 1st layer (low-level) and 5th layer (high-level) of VGG-M\cite{Chatfield2014} as the tracking features. 
\cite{Qi2016} uses 
the features extracted on different layers to build correlation filters and then combine 
these weak trackers into a strong one for tracking.  In these three methods, CNNs are solely treated as a feature extractor,
and they all adopt correlation filter as their base tracker. In addition, 
\cite{Hong2015} \ytyn{utilizes} Support Vector Machine (SVM) to model the target appearance based on the extracted CNN features. 

Because the labeled training data for tracking is limited, online training a convolution neural networks is prone to overfitting, which makes it becomes a challenging task. 
\cite{Wang2015} first \ytyn{pre-trains} a neural network to distinguish 
objects from non-objects on ImageNet \cite{Krizhevsky2012}, and then updates this model online with a differentially-paced fine-tuning scheme. A recent proposed tracker 
\cite{Nam2016} trained a multi-domain network, which has \abc{shared CNN layers to capture a generic feature represetnation}, and separate branches of subsequent domain-specific layers to do the binary classification (target vs. background) for each sequence.
\abc{For each sequence, the domain-specific layers must be fine-tuned to learn the target.}
\abc{In contrast to these CNN based methods that require running back-propagation to online train the network during tracking, our approach does not require online training, and instead uses a recurrent network to update the target appearance model with each frame. As a result, our tracker is faster than the CNN-based trackers that use online training.}

\abc{An alternative approach to target classification is to}
train a  similarity function for pairs of images, which regards visual tracking as an instance searching problem, i.e. using the target image patch on first frame as query image to search the object in the subsequent frames. 
Specifically, 
\cite{Tao2016} adopts a Siamese neural networks for tracking, which is a two-stream architecture, originally used for signature and face verification \cite{chopra2005}, and stereo matching \cite{zbontar2015}. 
\cite{Held2016} proposed a similar framework called GOTURN, which is trained to regress the target's position and size directly by inputting the network with a search image (current frame) and a query image (previous frame) that contains the target. 
\cite{Bertinetto2016} 
introduced a fully-convolutional Siamese network for visual tracking, which maps an exemplar of the target and a larger search area of second frame to a response map. 
\abc{In contrast to these methods,}
which do not have an online updating scheme that adapts the tracker to variations in the target appearance,
 our approach take the advantage of the LSTM's ability to capture and remember the target's appearance.

\subsection{RNN-based Trackers}

Recurrent Neural Networks (RNNs), especially its variants Long Short-Term Memory (LSTM) and Gated Recurrent Unit (GRU) are widely used in applications involve sequential data, such as language modeling \cite{mikolov2010recurrent}, machine translation \cite{sutskever2014sequence}, handwriting generation and recognition \cite{graves2013generating}, video analysis \cite{donahue2015long} and image captioning \cite{vinyals2015show}. 
Several recent works propose to 
solve the visual tracking problem as a sequential position prediction process by training an RNN to model the temporal relationship among 
frames. 
\cite{Gan2015} adopts a GRU to model the features extracted from a CNN,  while 
\cite{Kahou2016} embeds an attention mechanism into the RNN 
to guide 
target searching. However these two methods mainly focus on conducting experiments on synthesized data and have not demonstrate competitive results on commonly used tracking benchmark like OTB \cite{Wu2015} and VOT \cite{kristan2015visual}. 
\cite{Ning2016} employs the recent efficient and effective object detection method, 
YOLO \cite{Redmon2016}, to infer the 
preliminary position, 
and then use 
LSTM to estimate the target's bounding box. A heat map, 
constructed using the detected box from YOLO, is also used to help the RNN 
to distinguish object 
from non-object region. However, they only choose 30 sequences from OTB dataset for experiments, among which some of them are used as training data while other are used for testing. 

\yty{The major limitation of the aforementioned methods is that they directly feed the output of the fully connected layers into the original RNN or its variants (LSTM or GRU), which obfuscates the spatial information of target. 
\abc{In contrast, our approach better preserves the target's spatial structure, by changing the matrix multiplication in the fully-connected layers of the LSTM into convolution layers on feature maps.
As a result, our approach better preserves the spatial information of the target, while also being more memory-efficient.}
Furthermore, 
most existing RNN-based trackers estimate the target's bounding box by directly regressing it from the RNN hidden state, 
which is 
not effective because it requires 
enumerating all possible locations and scales of the target during training.} 
\abc{In contrast, our approach is based on estimating a response map of the target, which makes the training simpler and more effective.}

In contrast to modeling 
the temporal information of sequences as in \cite{Gan2015, Kahou2016, Ning2016}, 
\cite{Cui2016} uses 
RNN to capture the object's intrinsic structure by traversing a candidate region of interest from different directions. 
%
\cite{Fan2016} adopts similar strategy to capture spatial information of the target and incorporates it into MDNet \cite{Nam2016} to improve its robustness. 
In addition, 
a few recent works on multiple person detection \cite{Stewart2016} and tracking \cite{Milan2016} 
use RNNs to model the sequential relationship among different targets. 

\begin{figure}
	\begin{center}
		\includegraphics[width=1.05\linewidth]{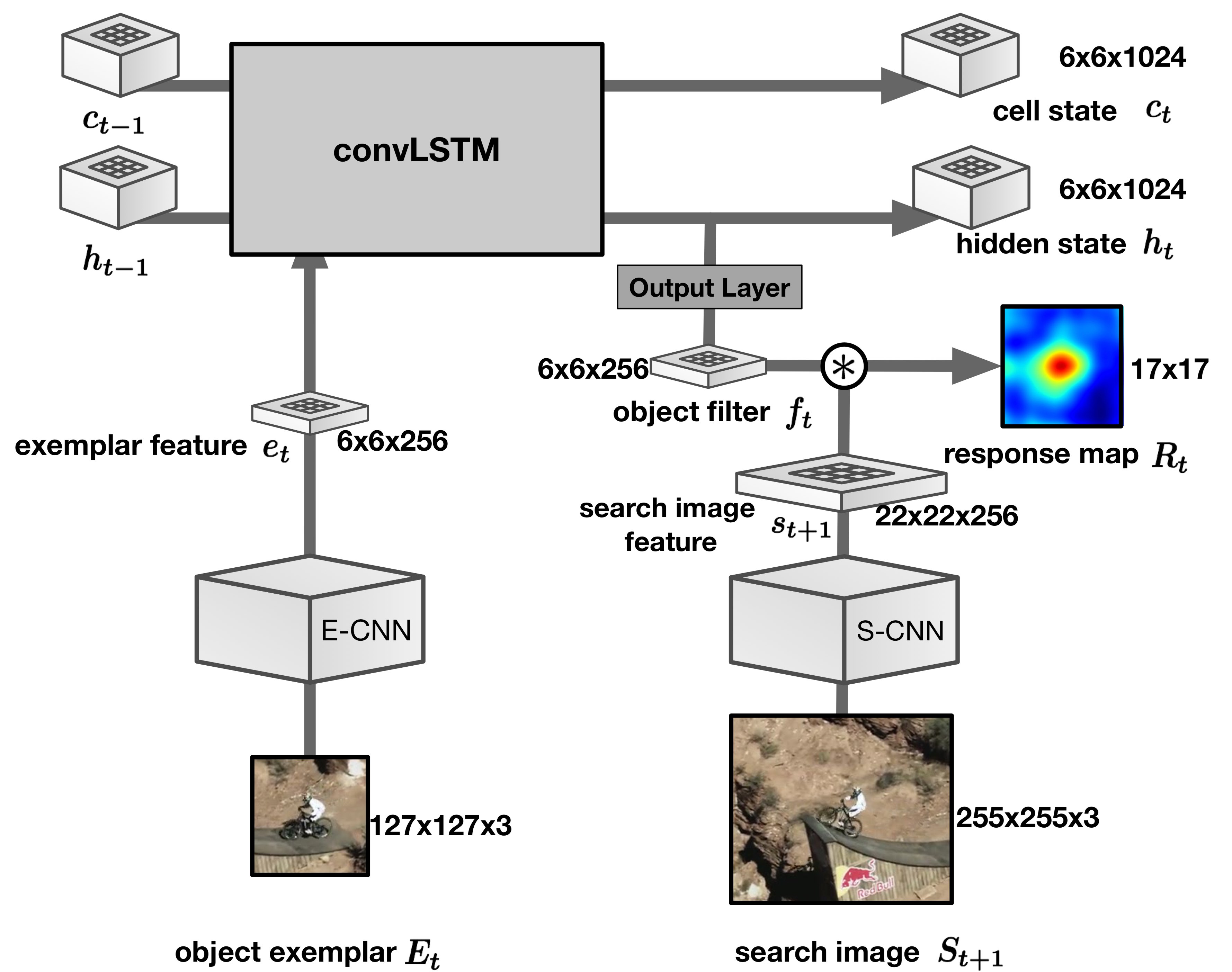}
	\end{center}
	\caption{The architecture of our recurrent filter learning networks.
		\abc{
			At time step $t$, a CNN (E-CNN) extracts features from the exemplar image patch.
			Using the previous hidden and cell states, $h_{t-1}$ and $c_{t-1}$, as well as the current \ytyn{exemplar} feature map $e_t$, 
			the convolutional LSTM memorizes the appearance information of the target by updating its cell and hidden states, $c_t$ and $h_t$.
			%
			The target object filter $f_t$ is generated by passing the new hidden state $h_t$ through an output convolutional layer.
			A feature map $s_{t+1}$ is extracted from the searching image (next frame) using another CNN (S-CNN), which is convolved by the target object filter $f_t$, resulting in a response map that is used to locate the target.}
	}
	\label{fig:1}
\end{figure}

\section{Recurrent Filter Learning}
This section describes our proposed filter generation framework for visual tracking. 

\subsection{Network Architecture}
\label{network}
The architecture of our networks is shown in Figure \ref{fig:1}. We design a recurrent fully convolutional network \abc{that extracts features from the object exemplar image and then generates a correlation filter, which is convolved with features extracted from the search image, resulting in a response map that indicates the position of the target.}
\abc{The network uses a convolutional LSTM to store the target appearance information from previous frames.}


 \yty{Our framework uses two CNN feature extractors: 1) the E-CNN is used to capture an intermediate representation of the object exemplar
\abc{for generating the target filter}; 2) 
the S-CNN is used to extract a feature map from the search image, which is convolved with the generated filter to produce the response map.
The E-CNN and S-CNN have the same architecture with different input image size, 127x127x3 and 255x255x3, respectively. 
As stated in \cite{Nam2016}, relatively smaller CNN are more appropriate for visual tracking because deeper and larger CNNs are prone to overfitting and dilute the spatial information. 
Therefore, for the two CNN feature extractors, 
we adopt a similar network as in \cite{Bertinetto2016}, which has five convolutional layers (see Table \ref{tab:1}).
%
Note that we do not share the parameters between these two CNNs because 
\abc{the optimal features used to generate the target filter could be different from the features needed to discriminate the target from the background}.
%
Indeed experiments in Section \ref{varcomp} show that sharing the CNNs' parameters decreases the performance.}

\begin{figure}
	\begin{center}
		\includegraphics[width=\linewidth]{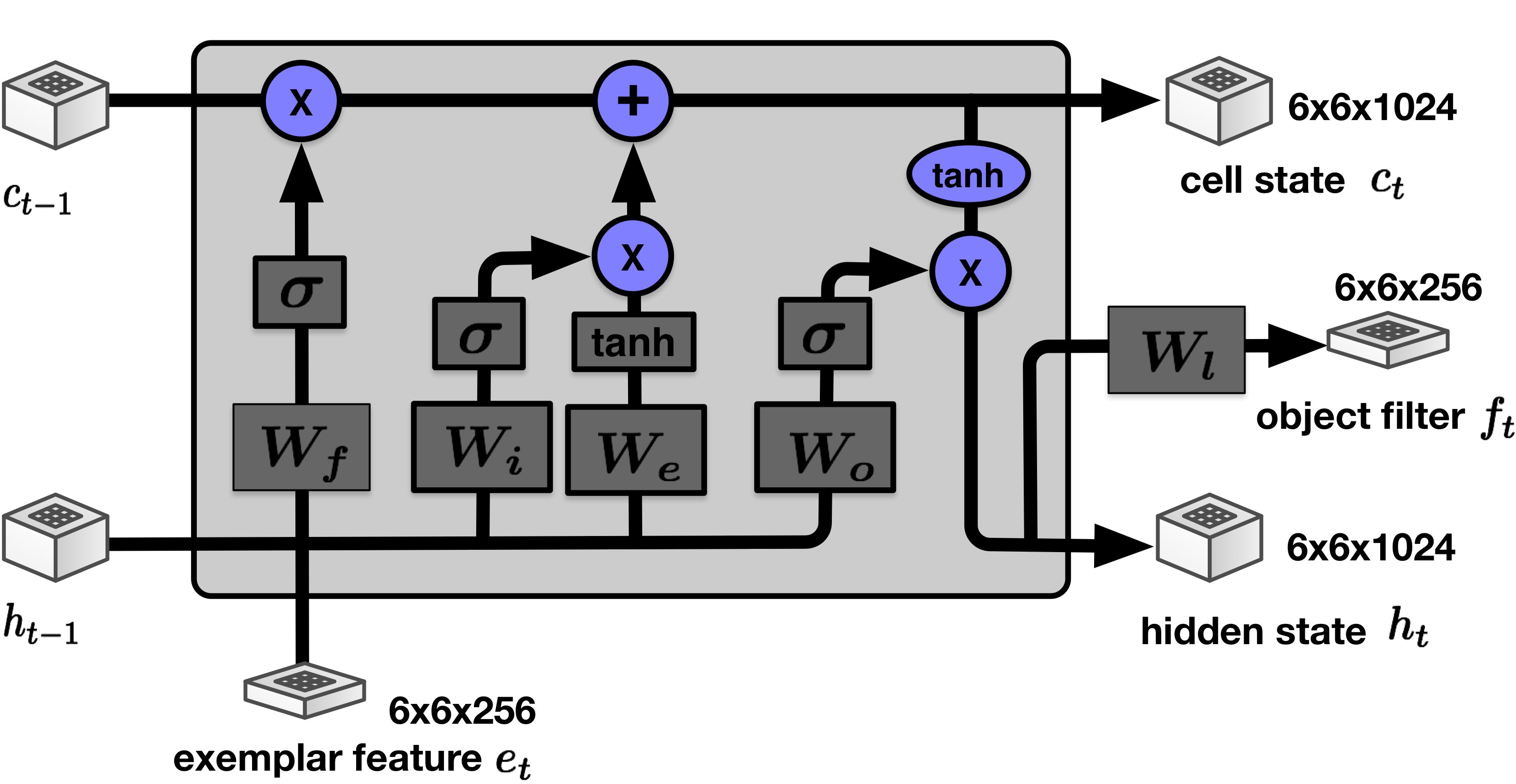}
	\end{center}
	\caption{\ytyn{The architecture of convolutional LSTM. $\sigma$ represents the sigmoid activation function and \textbf{tanh} denotes the tanh activation function. $W_f, W_i, W_e,W_o$ are the convolutional filter weights for the forget gate, input gate, estimated cell state, and output gate. All these convolutional filters have the same size 3x3x1024. $W_l$ is the convolutional weight of output layer with size 1x1x256.}}
	\label{Fig:con_lstm}
\end{figure}

\begin{table}[tbhp]
	\small
	\begin{center}
		\begin{tabular}{c|ccc}
			\hline 
			& filter size & channels & stride \\ 
			\hline 
			conv1 & 11x11 & 96 & 2 \\
			pool1 & 3x3 & - & 2 \\ 
			conv2 & 5x5 & 256 & 1 \\ 
			pool2 & 3x3 & - & 2 \\ 
			conv3 & 3x3 & 384 & 1 \\ 
			conv4 & 3x3 & 384 & 1 \\ 
			conv5 & 3x3 & 256 & 1 \\ 
			\hline
		\end{tabular} 
	\end{center}
	\caption{Architecture of the two CNN feature extractors.}
	\label{tab:1}
\end{table}

\abc{The architecture of the convolutional LSTM is shown in Figure \ref{Fig:con_lstm}.
The hidden $h_t$ and cell states $c_t$ are feature maps that are each 6x6x1024.}
The output of the E-CNN is a 6x6x256 convolutional feature map $e_t$, which is actually a grid of 256 dimensional feature vectors. 
Within the conv-LSTM, the exemplar feature map $e_t$ is concatenated to the previous hidden state $h_{t-1}$ along the channel dimension.
%
\abc{The various gates and update layers of the LSTM are implemented using convolutional layers with 3x3 filters, which helps to capture spatial relationships between features.}
Note that zero-padding is used in the convolution layers so that 
the hidden state and cell state have the same size with exemplar feature map.
%
\abc{The target filter $f_t$ is generated via the output layer of the LSTM, which is a convolution layer with a 1x1 filter that transforms the hidden state from 6x6x1024 to 6x6x256.}


Finally, to find the target in the next frame, the search image patch is input to the S-CNN to extract the search image feature map $s_{t+1}$, which is then convolved with the generated filter $f_t$ to produce a target response map.
%
%
Batch normalization \cite{Ioffe2015} is employed after each linear convolution to accelerating the converge of our network.  \ytyn{Each convolutional layers except conv5 is followed by \textit{Rectified linear units (ReLu)}. \textit{Sigmoid} functions are used after all gate convolution layers, and \textit{tanh} activation functions are used after the convolution layer for the estimated cell state.
There is no activation function for the output layer of the LSTM.}


\begin{figure}
	\begin{center}
		\includegraphics[width=\linewidth]{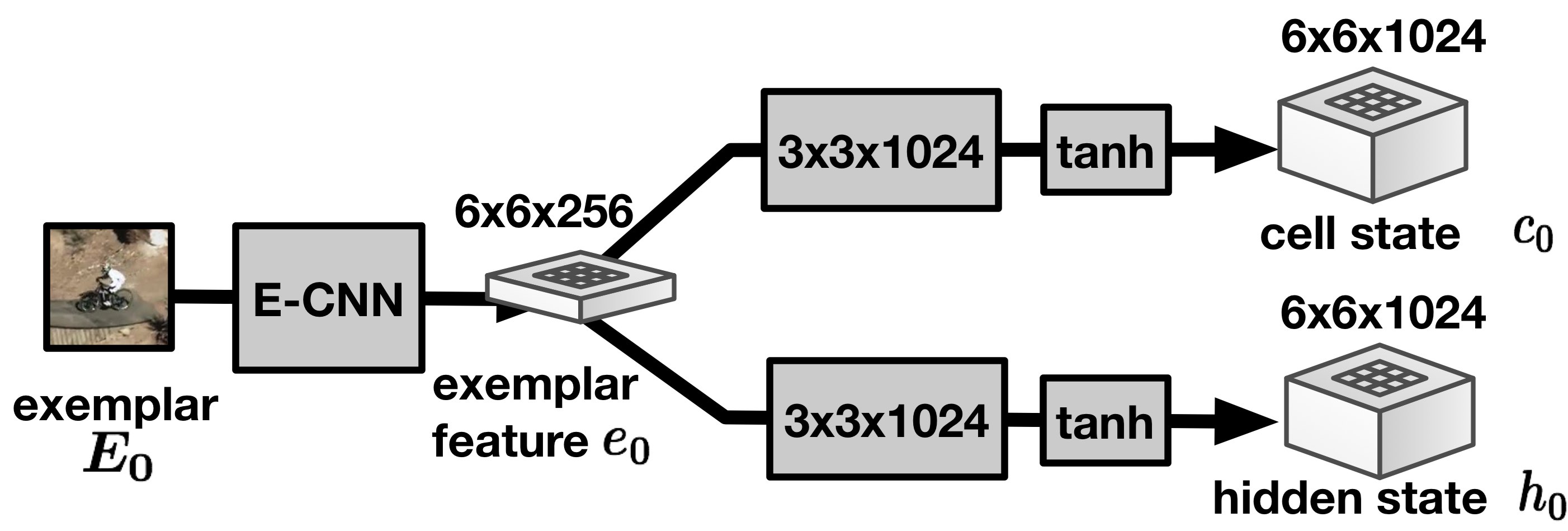}
	\end{center}
	\caption{\ytyn{Initialization network for memory units of the convolutional LSTM.}}
	\label{Fig:init}
\end{figure}

\subsection{Initialization of Memory States}
To start the recurrent filter learning process, we need to first initialize the memory state of the convLSTM using the exemplar target from the first frame. 
The initial exemplar target $E_0$ is input into the E-CNN to generate an exemplar feature map $e_0$, which is then fed into 
a convolutional layer to produce the initial hidden state $h_0$ (see Figure \ref{Fig:init}). The convolutional layer has filter size of 3x3, and output channel size of 1024.
A similar architecture is used to produce the initial cell state $c_0$ from the exemplar feature map $e_0$.
Here we use a tanh activation function after the convolution.
%
Although the numerical value of cell state may be out of the range [-1,1] (the output range of tanh) due to the addition operation between the input state and previous state, we experimentally find that this initialization method captures the information of initial target well. \ytyn{Our initialization network obtains about $10\%$ performance gain compared with initializing the cell state to zeros (see Section \ref{varcomp}).} 


\subsection{Loss Function}

The response map represents the probability of the target position on a 17x17 grid 
placed evenly around the center square of search image, with stride of 8. 
A loss function is applied on the response map to train the network. 
\abc{In particular, we generate a ground-truth response map, which consists of binary values that are computed from 
 the overlap ratio between the ground-truth bounding box and a bounding box centered at each position,
\begin{equation}
	\label{Eq:1}
	y_i = \begin{cases}
	1, & \text{if } \Phi(P_i, G) > \alpha\\
	0, & \text{otherwise}
	\end{cases}
\end{equation}
where, $P_i$ is a virtual bounding box centered on the i-th position of the response map, 
$G$ is the ground-truth bounding box, 
$\Phi(P, G)=\frac{P \cap G}{P \cup G}$ 
is the Intersection-over-Union (IoU) between regions $P$ and $G$, and $\alpha=0.7$ is the threshold.}

The loss between the predicted response map and ground-truth reponse map is the sum of the element-wise sigmoid cross-entropies,
\begin{equation}
	\label{Eq:2}
	L(p, y) = \sum_{i}-(1-y_{i})\log(1-p_{i})-y_{i}\log(p_{i}),
\end{equation}
where $p$ is the predicted response map, and $y$ is the ground-truth.

\subsection{Data and Training}
\label{sec:3.4}

ImageNet Large Scale Visual Recognition Challenge (ILSVRC) \cite{russakovsky2015imagenet} recently released a video data for evaluating  object detection from video. This dataset contains more than 4000 
sequences, where $\sim$3800 are used for training and $\sim$500 for validation. The dataset contains 
1.3 million labeled frames, and 
nearly two million object image patches marked with bounding boxes. 
This large dataset can be safely used to learn neural networks without worrying about overfitting, compared with traditional tracking benchmarks like OTB \cite{Wu2015} and VOT \cite{kristan2015visual}. As is discussed in \cite{Bertinetto2016}, the fairness of using sequences from the same domain to train and test neural networks for tracking is controversial. We instead train our model from the ImageNet domain to show our network generalization ability.

\begin{figure}
	\begin{center}
		\includegraphics[width=0.95\linewidth]{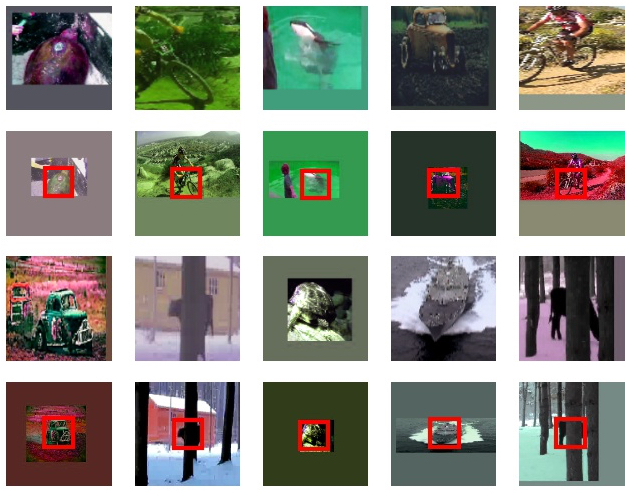}
	\end{center}
	\caption{Samples of target exemplar patches (1st and 3rd rows) and search images (2nd and 4th rows), from VID of ILSVRC. Targets are marked by a red bounding box on each search image.}
	\label{fig:2}
\end{figure}

The input of our framework is a sequence of cropped image patches. During training, we 
\abcn{uniformly sample}
$N$+1 
frames from each video.
\abc{Frames 1 through $N$ are used to generate the object \ytyn{exemplars}, and frames 2 to $N$+1 are the corresponding search image patches.}
Both the object exemplar and search image patches are cropped around the center of groundtruth bounding box.  The exemplar size is two times that of the target, and the search image size is four times the target size, as is shown in Figure \ref{fig:2}. 
The reason that the exemplar size is larger than the target size is 
 to cover some context information, which may provide some 
negative examples 
when generating the filter. 
If the patch extends 
beyond the image boundary,  
we pad using the mean RGB value of the image. 
\abc{Note that our architecture is fully convolutional, which is translation invariant.  Hence, it suffices to train the network using search images where the targets are always located at the center.}

\subsection{Training Details}
The recurrent model is trained by minimizing the loss in Eq. \ref{Eq:2} using the ADAM \cite{kingma2014adam} method with a mini-batches of 10 video clips of length 10. 
The initial learning rate is 1e-4, which is decayed by a multiplier 0.8 every 5000 iterations. In case of gradient explosion, gradients are clipped \ytyn{to a maximum magnitude of 10}. 

The exemplar and search image patches are pre-processed by resizing them to their corresponding CNN input sizes.
We use various forms of data augmentation.  Following \cite{Krizhevsky2012}, we randomly distort colors by randomly adjusting image brightness, contrast, saturation and hue. 
In addition, we apply 
minor image translation and stretching to both the exemplar and search images. 
%
Search images are shifted around the center with a range of 4 pixels (within the stride of the response map grid).
The target in exemplar patches can be translated anywhere as long as its bounding box is within the exemplar patch.
%
For stretching, we randomly crop the image patches, 
and resize them into their network input size. 
We set a 
small maximum stretch range (0.05 times of the target size) for search images, and larger stretch range (0.5 times of the target size) for exemplar patch. 
This setting accounts for the 
situations where 
the estimated target position and scale 
may not be very accurate, and hence  
the target in the 
exemplar is shifted from the center or stretched due to resizing. 
%
The first frame of the clip for exemplar has no data augmentation since we assume that 
the initial bounding box is always accurate in visual tracking.  Although the object \ytyn{in the exemplar patch} may appear in various places within 
the patch because of augmentation, the memory unit of our model can still capture the information of target due to its convolutional structure.

\section{Online Tracking}

We use a simple strategy for online tracking with our RFL framework, which is presented 
in Algorithm \ref{algo:1}. Unlike other CNN-based 
algorithms \cite{Nam2016, Nam20161, Wang2015}, we do not online fine-tune the parameters of our networks using SGD back-propagation, \abc{as the target appearance is represented by the hidden and cell states of the conv-LSTM.}
Furthermore, we do not refine our predictions using bounding box regression as in \cite{Nam2016, Tao2016, Fan2016}. 
Instead, we directly upsample the response map into a larger size using bicubic interpolation as in \cite{Bertinetto2016}, and choose the position with maximum value as the target's location. 
To account 
for 
scale changes, we calculate response maps at different scales by building a pyramid of scaled search images, resizing them to the same input size, and then assembling them into a mini-batch to be processed by the network in a single feed-forward pass.


During tracking, we first use the target exemplar to initialize the memory state of LSTM using the initialization network (Figure \ref{Fig:init}). For the subsequent frames, we convolve the filter generated by the conv-LSTM with the search images, which are centered at the previous predicted position of object, to get the response map. 
Let $\{S^{-1}_t, S^0_t, S^1_t\}$ be three scales of the search image.  Using the S-CNN, we obtain the 
corresponding feature maps $\{s^{-1}_t, s^0_t, s^1_t\}$.  The response map for scale $m$ is calculated as,
\begin{equation}
\label{Eq:3}
	\ytyn{R^m(f_{t}, s_{t+1}) = f_{t}\ast s^m_{t+1}}.
\end{equation}
Let $v^m$ be the maximum score of the response map $R^m$ at scale $m$.
Then, the predicted target scale is the scale with largest maximum score,
\begin{equation}
\label{Eq:4}
	m_{best} = \argmax_{m}{v^m}
\end{equation}
The predicted target position is obtained \ytyn{by} averaging positions with the top $K$ response values on the score map $R^{m_{best}}$ of the predicted scale,
\begin{equation} 
\label{Eq:5}
	p^* = \ytyn{\frac{1}{K}}\sum^K_{k} p_{k}
\end{equation}
where $p_{k}$ are the top $K$ locations.


\begin{algorithm}[htb]   
	\caption{Online Tracking Algorithm with RFL}
	\label{algo:1}   
	\begin{algorithmic}[1] 
		\REQUIRE ~~\\
		Pretrained RFL model $M$;\\  
		Initial target exemplar $E_0$;\\    
		\ENSURE ~~ \\
		Predicted object's bounding box\\
		\STATE Initialize the memory state of conv-LSTM  using the initialization network on exemplar $E_0$;
		\STATE Generate an initial filter $f_0$ using the conv-LSTM. 
		\FOR{$t=\{0,\cdots,T-1\}$}
		\STATE Get search image $S_{t+1}$ from frame $t+1$.
		\STATE Build a pyramid of scaled search images $\{S^{-1}_{t+1}, S^0_{t+1}, S^1_{t+1}\}$, and apply S-CNN on each scaled image to produce feature maps $\{s^{-1}_{t+1}, s^0_{t+1}, s^1_{t+1}\}$.
		\STATE Convolve the filter $f_{t}$ with $\{s^{-1}_{t+1}, s^0_{t+1}, s^1_{t+1}\}$ as in Eq.~\ref{Eq:3}.
		\STATE Upsample the responses using bicubic interpolation and penalize scale changes.
		\STATE Normalize and apply cosine window on score maps $\{R^{-1}_{t+1}, R^0_{t+1}, R^1_{t+1}\}$.
		\STATE Estimate the target's scale by Eq.~\ref{Eq:4}.
		\STATE Predict the target's position by Eq.~\ref{Eq:5}.
		\STATE Crop a new target exemplar $E_{t+1}$ centered at the estimated position in frame $t+1$.
		\STATE Update the conv-LSTM using exemplar $E_{t+1}$. 
		\STATE Generate a new filter $f_{t+1}$ from the conv-LSTM.
		\ENDFOR
	\end{algorithmic}  
\end{algorithm}

After we get the estimated position, a target exemplar is cropped from the frame, and used to update the conv-LSTM. 
As seen in recent works \cite{Tao2016, Danelljan2016}, excessive model updating is prone to overfitting to recent frames.
Hence, we dampen the memory states of the conv-LSTM using \abc{exponential decay smoothing}, $A_{est} = (1-\beta)*A_{old}+\beta A_{new}$, where $\beta$ is the decay rate and $A$ is the memory state.
%
Furthermore, we also penalize the score map of other scales (scales $\pm1$) by multiplying those score maps by a penalty factor $\gamma$.
To smoothen the scale estimation and penalize large displacements, we use exponential decay smoothing. 
%


\subsection{Tracking Parameters}
We adopt three scales $1.03^{[-1,0,1]}$ to build the scale pyramid of search images.
The penalty factor on score maps of non-original scales is $\gamma=0.97$.
Object scale is damped with decay rate of $\beta=0.6$, while memory states are damped with a decay rate of $\beta=0.06$. Score map is damped with a cosine window by the decay rate of $\beta=0.11$. \ytyn{What' more, the number of candidate position used for averaging final prediction is $K=5$.}

\section{Experiments}

We evaluate our proposed RFL tracker on 
OTB100 \cite{Wu2015}, VOT2016 \cite{Kristan2016} and VOT2017. We implement our algorithm in Python using the TensorFlow toolbox \cite{abadi2016tensorflow}, and test it on a computer with four Intel(R) Core(TM) i7-6700 CPU @ 3.40GHz and a single NVIDIA GTX 1080 with 8G RAM.  The running speed of our algorithm is about 15 fps.

\subsection{Tracking Results on OTB100}
There are two common performance metrics for evaluating tracking, 
center location error (CLE) and Intersection-over-Union (IoU). 
Specifically, CLE measures the pixel distance error between the predicted and groundtruth \abc{center} positions.
IoU measures the overlap ratio between the predicted and ground-truth bounding boxes. However, as discussed in \cite{Wu2015}, CLE only evaluates the pixel difference, which is actually not fair for small objects, and does not consider the scale and size changes of objects. We thus use IoU success plots as our evaluation metric on OTB.

\begin{figure}[tbhp]
	\begin{center}
		\includegraphics[width=0.49\linewidth]{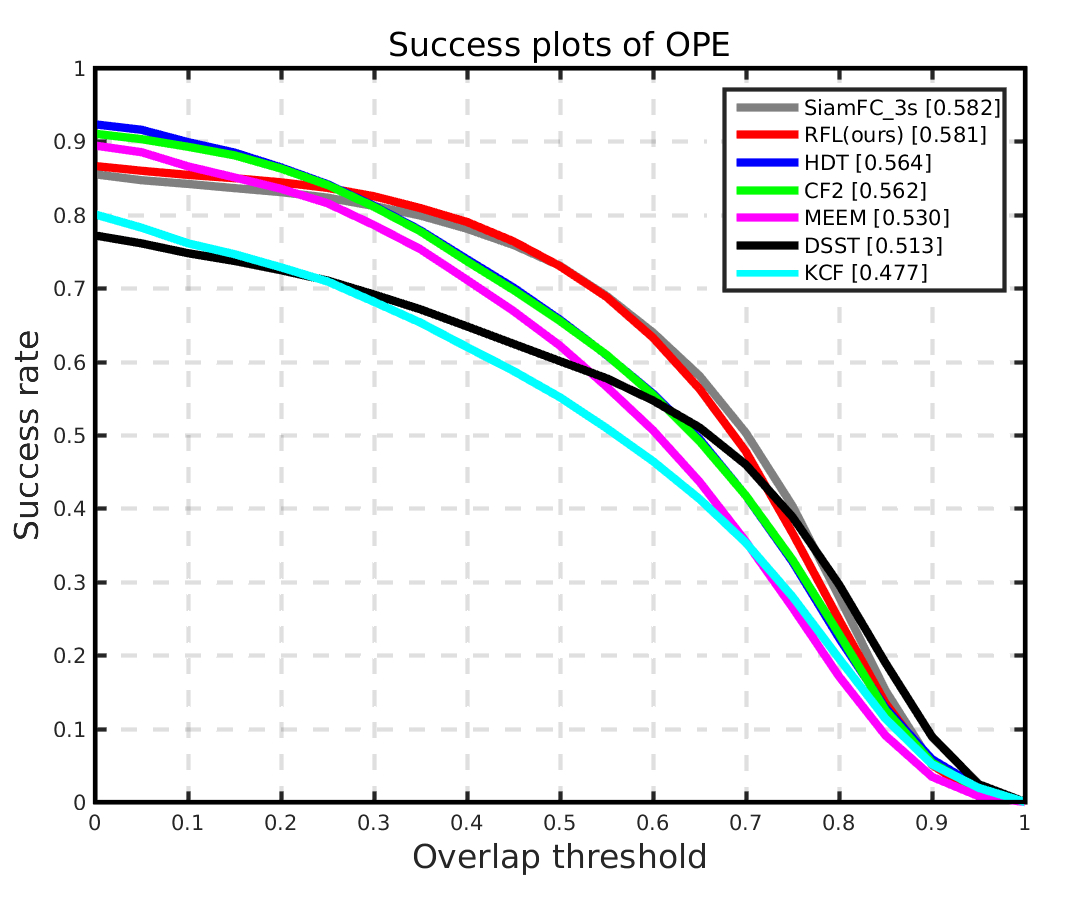}
		\includegraphics[width=0.49\linewidth]{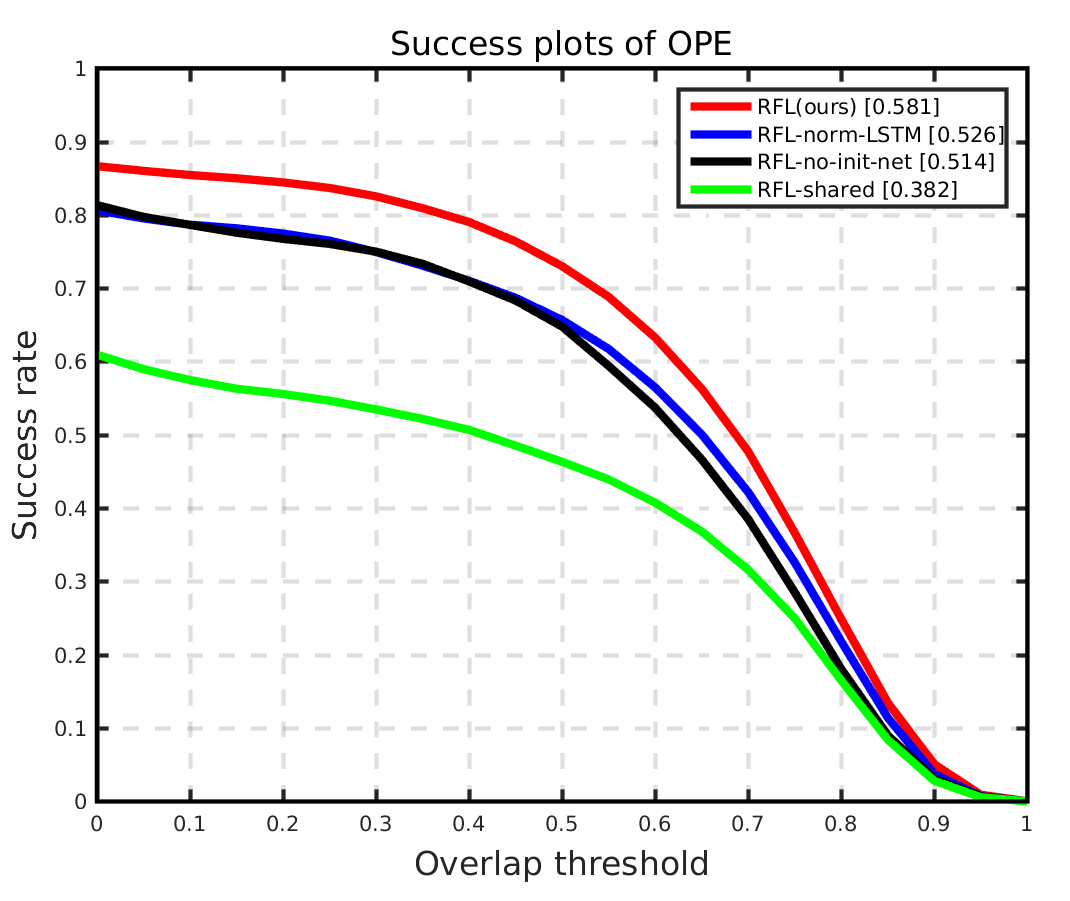}
	\end{center}
	\caption{Left: Success plots of OPE on OTB100 compared with seven trackers. Right: Success plot of OPE on OTB100 for different variants of proposed RFL. Trackers are ranked by their AUC scores in the legend. Our proposed method is RFL.
	}
	\label{Fig:3}
\end{figure}

\begin{figure*}[tbhp]
	\begin{center}
		\includegraphics[width=0.85\linewidth]{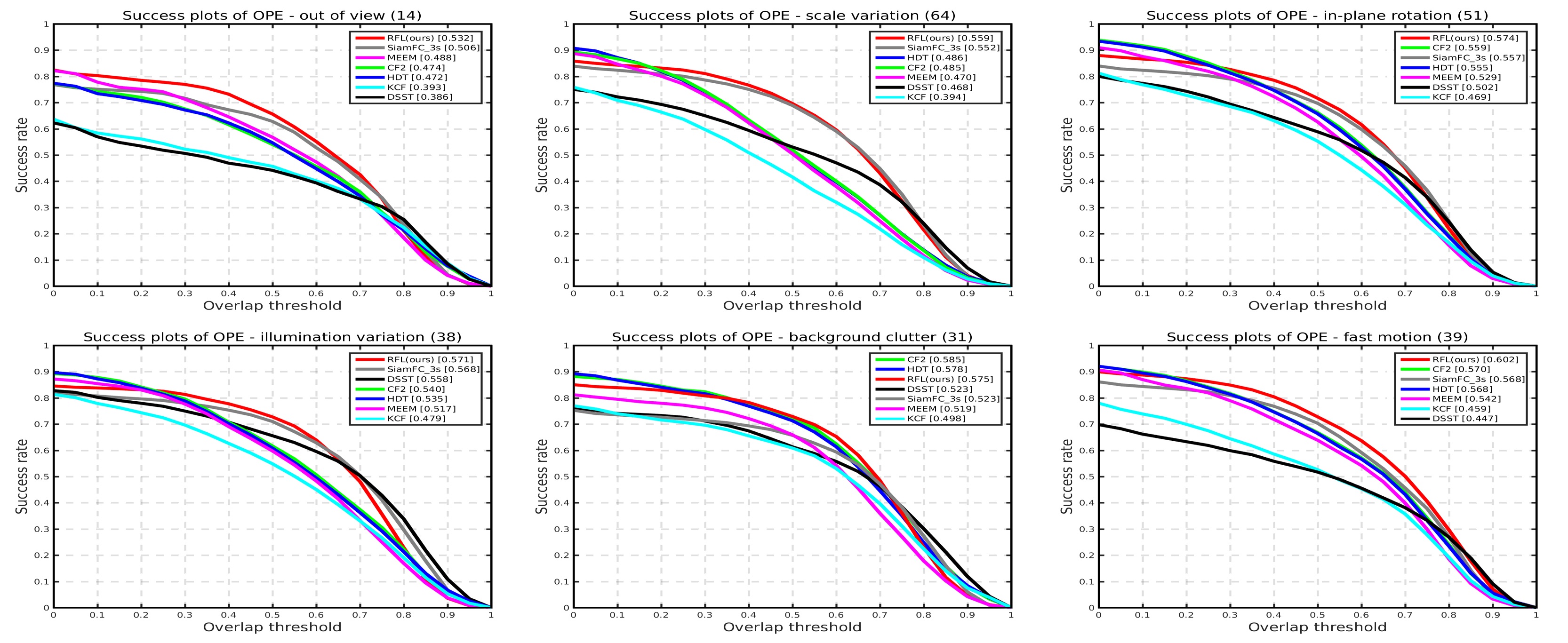}
	\end{center}
	\caption{Success plot for 6 challenge attributes: out-of-view, scale variation, in-plane rotation, illumination variation, background clutter and fast motion. Trackers are ranked by their AUC scores in the legend.}
	\label{Fig:4}
\end{figure*}

 We compare our tracking method with six other state-of-the-art methods, including SiamFC \cite{Bertinetto2016}, CF2 \cite{Ma2015}, HDT \cite{Qi2016}, MEEM \cite{Zhang2014}, DSST \cite{Danelljan2014}, and KCF \cite{Henriques2015} on OTB100. We present the OPE success plots of  in Figure \ref{Fig:3} (Left), where our method is denoted as RFL. 
Trackers are ranked by their \textit{area under curve} (AUC) score.
Overall, SiamFC has slightly higher AUC (ours .581 vs .582). Looking closer (see Figure \ref{Fig:4}), our RFL has higher AUC on videos where the target or background appearance changes compared with SiamFC: fast motion (ours .602 vs .568), background clutter (ours .575 vs .523), out-of-view (ours .532 vs .506). I.e. our RFL can better adapt the tracking filter over time to better discriminate the changing target/background.

We also report the average success rates at various thresholds 
in Table \ref{tab:2}. In addition to our superiority in AUC score, our method's success rates at IoU=0.3 and IoU=0.5 is much higher than other methods or on par with SiamFC, 
which suggests 
that our RFL predicts a more accurate bounding box than other trackers.

\begin{table}
\small
	\begin{center}
		\begin{tabular}{c|c|ccc}
			\hline 
			Tracker & AUC & P@0.3 & P@0.5 & P@0.7 \\ 
			\hline 
			RFL & 0.581 &  \textbf{0.825} & \textbf{0.730} & \underline{0.477} \\
			
			SiamFC\_3s & \textbf{0.582} & \underline{0.812} & \textbf{0.730} & \textbf{0.503} \\ 
			HDT & 0.564 & \underline{0.812} & \underline{0.657} & 0.417 \\ 
			CF2 & 0.562 & 0.811 & 0.655 & 0.417 \\ 
			MEEM & 0.530 & 0.786 & 0.622 & 0.355 \\ 
			DSST & 0.513 & 0.691 & 0.601 & 0.460 \\ 
			KCF & 0.477 & 0.681 & 0.551 & 0.353 \\ 
			\hline 
		\end{tabular} 
	\end{center}
	\caption{AUC and success rates at different IoU thresholds.}
	\label{tab:2}
\end{table}

\begin{figure*}
	\begin{center}
		\includegraphics[width=0.8\linewidth]{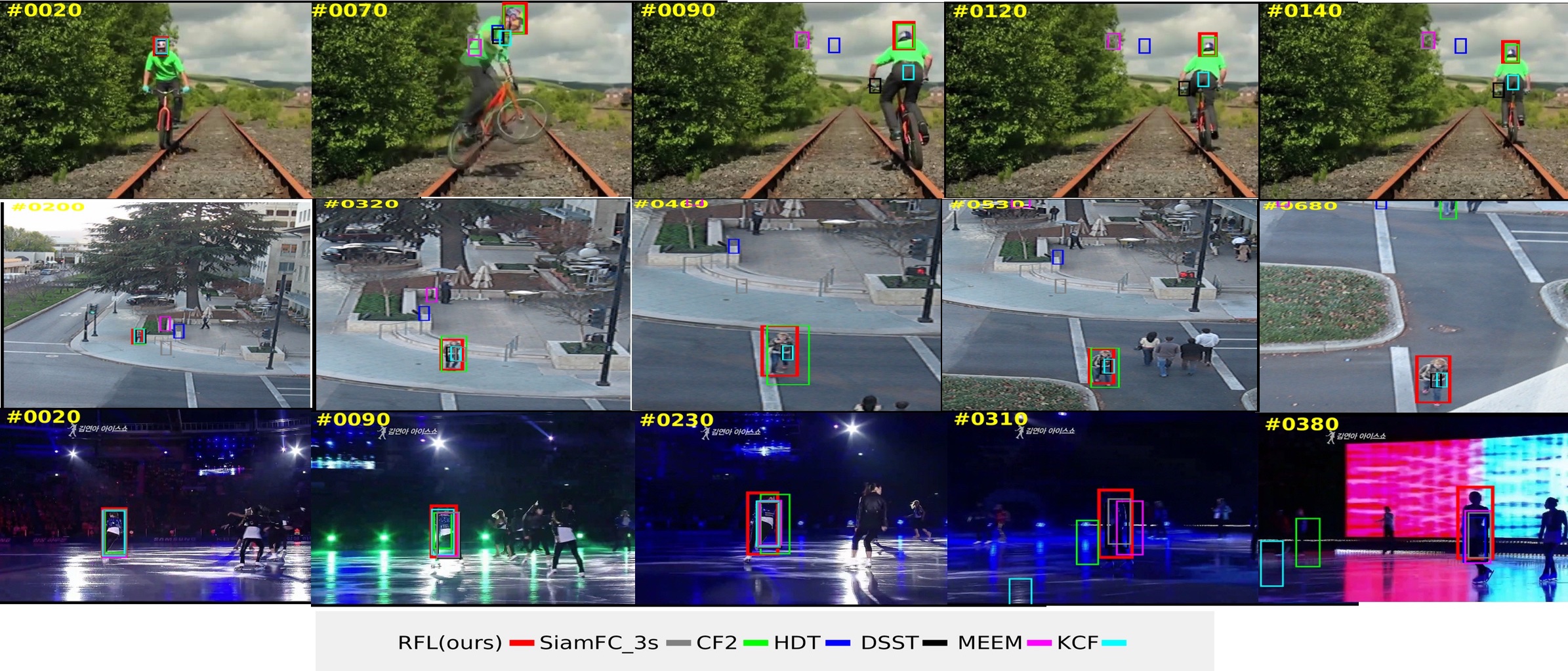}
	\end{center}
	\caption{Qualitative results of our method on some challenging videos (from top to botton, \textit{biker, human5, skating1})}
	\label{Fig:5}
\end{figure*}

We further analyze the performance of our method under different sequence attributes annotated in OTB. 
Figure \ref{Fig:4} shows the OPE results comparison of 6 challenge attributes: \textit{out-of-view, scale variation, in-plane rotation, illumination variation, background clutter and fast motion}. 
%
Our RFL tracker shows better or similar 
performance compared with other methods. 
Note that our method works especially well under the \textit{fast motion}  and \textit{out of view} attribute. Figure \ref{Fig:5} shows some example frames of the results generated by our RFL and the other six trackers
on several challenging sequences \abc{(see supplemental for vidoes)}. From these videos, we can see that our method handles fast motion, scale changes and illumination variation better than others.


\subsection{Tracking Results on VOT2016 and VOT2017}
\begin{figure}
	\begin{center}
		\includegraphics[width=0.9\linewidth]{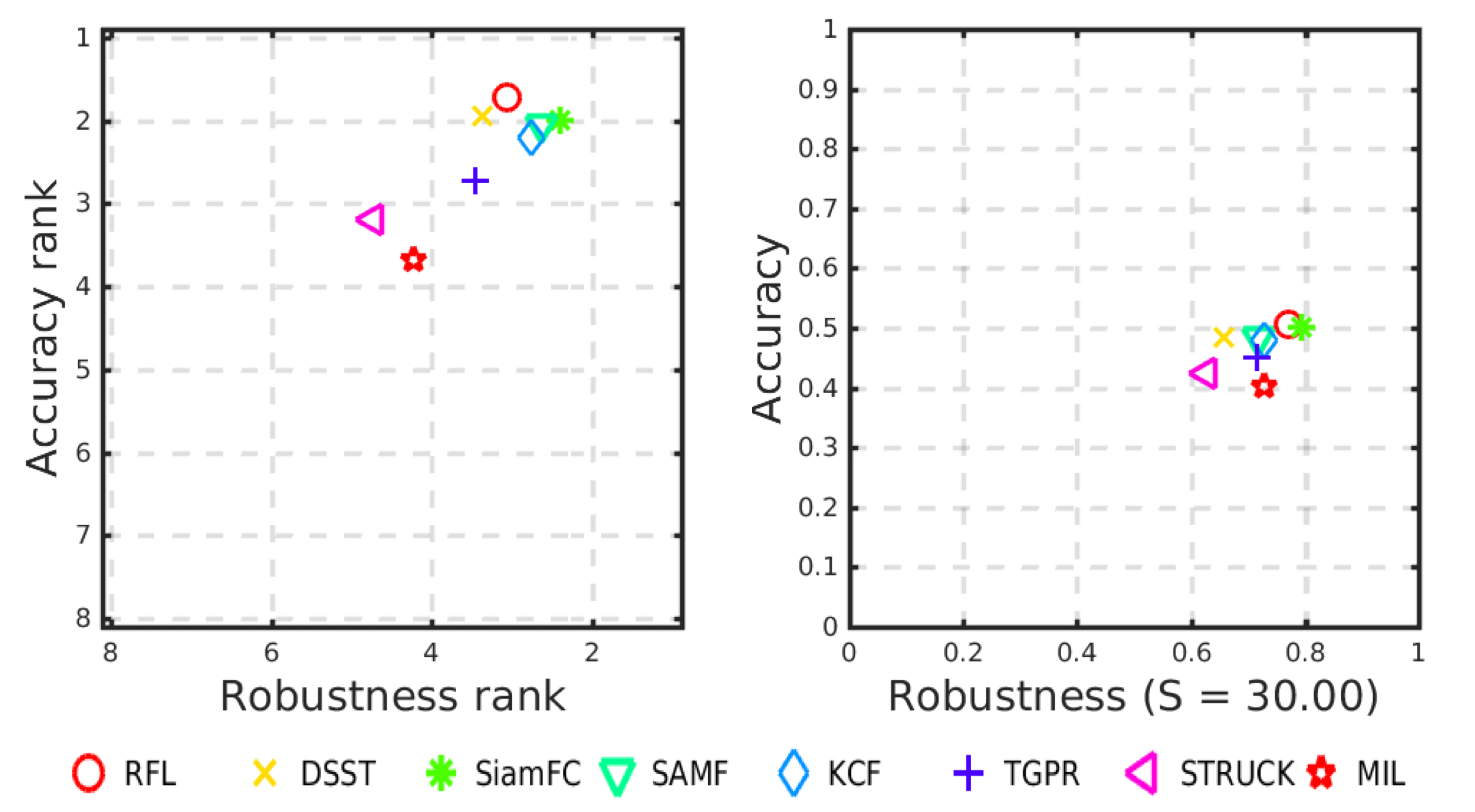}
	\end{center}
	\caption{A state-of-the-art comparison on the VOT2016 benchmark. Left: Ranking plot includes the accuracy and robustness rank for each tracker. Right: AR plot shows the accuracy and robustness scores}
	\label{Fig:6}
\end{figure}

We presents a comparison of our trackers with 7 state-
of-the-art trackers, including DSST \cite{Danelljan2014}, SiamFC \cite{Bertinetto2016}, SAMF \cite{Li2014}, KCF \cite{Henriques2015}，TGPR \cite{Gao2014}, STRUCK \cite{Hare2011} and MIL \cite{Babenko2011} on VOT2016 \cite{Kristan2016}. Two performance metrics, which are \textit{accuracy} and \textit{robustness}, are used to evaluate our algorithm. The \textit{accuracy} measures the averaged per-frame overlap between predicted bounding box and ground truth,  while \textit{robustness} reveals the number of failures over sequences. We generate these results using the VOT-toolkit. Figure \ref{Fig:6} visualizes independent ranks for each metric on two axes.  Table \ref{tab:3} reports the AR ranking (Accuracy and Robustness) and EAO (expected averaged overlap) for each tracker. Our RFL tracker achieves the best accuracy rank and shows competitive EAO compared with other trackers.

We also evaluate our method on VOT2017 challenge. In baseline experiment, we obtain an average overlap of 0.48 and failure of 3.29. In unsupervised experiment, we get an average overlap of 0.23. Our average speed in realtime experiment is 15.01 fps.

\begin{table}
	\begin{center}
		\begin{tabular}{c|c|c|c}
			\hline
		   	Tracker &  {Accuracy} & {Robustness} & {EAO} \\
			\hline
			{RFL} & \textbf{1.72} & 3.07 & \underline{0.2222}\\
			{DSST} & \underline{1.95} & 3.38 & 0.1814 \\
			{SiamFC\_3s} & {2.00} & \textbf{2.42} & \textbf{0.2352} \\
			{SAMF} & 2.05 & \underline{2.65} &  0.1856\\
			{KCF} & 2.20 & {2.77} &  0.1924\\
			{TGPR} & 2.73 & 3.47 & 0.1811 \\
			{STRUCK} & 3.20 & 4.73  & 0.1416 \\
			{MIL} & 3.68 & 4.23 & 0.1645 \\
			\hline
		\end{tabular}
	\end{center}
\caption{AR ranking and EAO for each tracker.}
\label{tab:3}
\end{table}

\subsection{Experiments on Architecture Variations}
\label{varcomp}
To show the effectiveness of our recurrent filter learning framework, we also conduct comparisons among three variations of our framework: 
\begin{compactitem}
	\item Using shared the weights for E-CNN and S-CNN (denoted as \textit{RFL-shared}).
	\item Initializing the memory states to all zeros, rather than using the intialization network 
	(denoted as \textit{RFL-no-init-net}). 
	\item Use a normal LSTM on each channel vector of the feature map, 
	i.e.\ by setting the filter sizes in the convolutional LSTM to 1x1 (denoted as \textit{RFL-norm-LSTM}) 
\end{compactitem}

Figure \ref{Fig:3} (Right) shows the success plot of OPE on OTB100 for different variants of RFL. 
Sharing parameters between the E-CNN and S-CNN significantly decreases the tracking performance.
Also, we are able to obtain about 8-10\% 
performance gain by using the initialization network for the memory state and a convolutional LSTM.



\section{Conclusion}

In this paper, we have explored the effectiveness of generating tracking filters recurrently by sequentially feeding target exemplars
into a convolutional LSTM for visual tracking.
To best of our knowledge, we are the first to demonstrate improved and competitive results on large-scale tracking datasets (OTB100, VOT2016 and VOT2017) using RNN to model temporal sequences. 
Instead of initializing and updating the neural network using time-consuming SGD back-propagation for each specific video as in other CNN-based methods \cite{Nam2016, Nam20161, Wang2015}, 
our tracker estimates the target filter using only feed-forward computation.
Extensive experiments on well-known tracking benchmarks have validated the efficacy of our RFL tracker.

\small
\section*{Acknowledgements}
This work was supported by the Research Grants Council of the Hong Kong Special Administrative Region, China (CityU 11200314), and by a Strategic Research Grant from City University of Hong Kong (Project No. 7004682). We gratefully acknowledge the support of NVIDIA Corporation with the donation of the Tesla K40 GPU used for this research.

{\small
\bibliographystyle{ieee}
\bibliography{egbib}
}

\end{document}